\documentclass[conference]{IEEEtran}
\IEEEoverridecommandlockouts

\usepackage{graphicx} 
\usepackage{transparent} 
\usepackage{xcolor}
\usepackage{subfigure}
\usepackage{epsfig} 
\usepackage{mathptmx} 
\usepackage{times} 
\usepackage{amsmath} 
\usepackage{amssymb}  
\usepackage{mathtools}
\usepackage{url}
\usepackage{calc}
\usepackage{tikz}
\usepackage{listings}
\usepackage{tabularx}
\usepackage{stix}
\usepackage{cite}
\usepackage{algorithm}%
\usepackage{algpseudocode}
\usepackage{balance}
\usetikzlibrary{shapes,arrows}

\usepackage[nolist]{acronym} 
\newacro{bt}[BT]{Behavior Tree}
\newacro{rbt}[RBT]{Reconfigurable Behavior Tree}

\graphicspath{{img/}}

\title{Reconfigurable Behavior Trees: Towards an Executive Framework Meeting High-level Decision Making and Control Layer Features\\
\thanks{This research has received funding from the European Union's Horizon 2020 research and innovation programme (grant agreement no. 731761, IMAGINE) and from the Austrian Research Foundation (Euregio IPN 86-N30, OLIVER)}
}

\author{\IEEEauthorblockN{Pilar de la Cruz}
\IEEEauthorblockA{\textit{Dept. of Computer Science} \\
\textit{University of Innsbruck}\\
Innsbruck, Austria \\
pilar.de-la-cruz@uibk.ac.at}
\and
\IEEEauthorblockN{Justus Piater}
\IEEEauthorblockA{\textit{Dept. of Computer Science} \\
\textit{and Digital Science Center} \\
\textit{University of Innsbruck}\\
Innsbruck, Austria \\
justus.piater@uibk.ac.at}
\and
\IEEEauthorblockN{Matteo Saveriano}
\IEEEauthorblockA{\textit{Dept. of Computer Science} \\
\textit{and Digital Science Center} \\
\textit{University of Innsbruck}\\
Innsbruck, Austria \\
matteo.saveriano@uibk.ac.at}
}

\definecolor{dgreen}{rgb}{0.0, 0.5, 0.0}

\tikzstyle{every node}=[sibling distance=5em, level distance=3em, anchor=center]
\tikzset{
  edge from parent/.style = {draw, ->, below, -latex},
  ControlNode/.style = {shape=rectangle,
  											minimum size=6mm,
  											thick,
                     		draw, align=center},
  Sequence/.style = {ControlNode, font={\normalsize$\rightarrow$}},
  Fallback/.style = {ControlNode, font={\normalsize$?$}},
  Parallel/.style = {ControlNode, font={\normalsize$\rightrightarrows$}},
  Decorator/.style = {shape=diamond, thick, draw, align=center},
  Action/.style = {shape=rectangle, inner sep=2, font={\scriptsize\tt}, draw, align=center},
  Condition/.style = {shape=ellipse, inner sep=1, font={\scriptsize\tt}, draw, align=center}
}

\begin{document}

\maketitle
\thispagestyle{empty}
\pagestyle{empty}

\begin{abstract}
\acp{bt} constitute a widespread artificial intelligence tool that has been successfully adopted in robotics. Their advantages include simplicity,  modularity, and reusability of code. However, Behavior Trees remain a high-level decision making engine; control features cannot easily be integrated. 
This paper proposes \acp{rbt}, an extension of the traditional \acp{bt} that incorporates sensed information coming from the robotic environment in the decision making process. We endow \acp{rbt} with continuous sensory data that permits the online monitoring of the task execution. The resulting stimulus-driven architecture is capable of dynamically handling changes in the executive context while keeping the execution time low. The proposed framework is evaluated on a set of robotic experiments. The results show that \acp{rbt} are a promising approach for robotic task representation, monitoring, and execution. 


\end{abstract}

\begin{IEEEkeywords}
Decision making, Task representation, Continuous execution monitoring
\end{IEEEkeywords}

\section{Introduction}\label{sec:intro}
Modern robotic agents demand increased flexibility in the way the assigned task is represented and executed. Indeed, autonomous robotic systems need to be capable of dealing with sensing and planning operations at low cost, as well as monitoring actions in a goal-oriented fashion~\cite{lopez2019formal}. Behavior Trees (BTs) constitute a powerful tool for task switching and decision making, and are receiving increasing attention in the robotics community~\cite{colledanchise2018behavior, colledanchise2016behavior, rovida2017extended}. The reason for this growing attention mostly lie in the fact that \acp{bt} are self-explanatory, modular, facilitate code reuse, and are simple to design. A \ac{bt} is built combining a limited number (six) of node types. This greatly simplifies the design of new \acp{bt}, makes them \textit{human-readable}, and eases the formal verification of the generated task plan without curbing their expressive power~\cite{colledanchise2016behavior}. {Modularity has been exploited in robotics in both hardware \cite{yim2007modular,moubarak2012modular} and software \cite{elkady2012robotics}.} \acp{bt} are modular in the sense that each subtree may be seen as a subblock that may be added or replaced by any other subblock. This makes the code reusable for different applications and further simplifies the design of new \acp{bt}. 


However, \ac{bt} engines are not designed to operate within the sense-plan-act paradigm, nor do they provide an optimized trade-off between reactiveness and execution cost for low-level control. Also, \acp{bt} may easily grow when the number of actions and conditions needed for closing the execution loop increases. 
Moreover, continuous monitoring of the task execution as well as online resolution of possible ambiguities in the task plan are typically not supported by \acp{bt}. {This limits the applicability of \acp{bt} in dynamic or uncertain environments.} In order to overcome these limitations, a robotic executive framework has to: \textit{i)} ensure low complexity in terms of cost and implementation when dealing with task executions, \textit{ii)} make a connection between low-level stimuli and high-level decision making, and \textit{iii)} perform a continuous and goal-oriented planning.
	

This paper proposes an executive framework that meets the robotic task requirements by combining the planning capabilitites of \acp{bt} with attentional mechanisms for control features~\cite{kawamura2008implementation, caccavale2016flexible}. The proposed \textit{Reconfigurable Behavior Trees (RBTs)} exploit the high modularity of traditional \acp{bt} to define a tree structure that can be reconfigured at runtime, i.e. dynamically during the task execution, by adding and/or removing parts of the tree. The reconfiguration is ruled by environmental stimuli corresponding to changes in the sensed information and by the successful execution of goal-directed actions. 
This paper presents a formal definition of the \ac{rbt} framework and evaluates its performance {in scheduling robotic tasks.} 
{To summarize, our contribution is two-fold:
\begin{itemize}
	\item We extend the \ac{bt} formalism to incorporate continuous information in the decision process. This makes it possible to monitor the execution and react to unexpected changes in the environment. 
	\item We propose an algorithm to dynamically allocate and deallocate tree branches. Subtrees are stored as JSON (JavaScript Object Notation) schemas, a format that is supported by most existing databases.
\end{itemize}
} 

Section~\ref{sec:related} presents related work. 
In Sec.~\ref{sec:rbt}, we describe the proposed approach. Section~\ref{sec:evaluation} presents simulation experiments and evaluates the results. Section~\ref{sec:conclusion} states the conclusions and proposes further extensions.

\section{Related Work}
\label{sec:related}

Finite state machines have been widely used in different areas of computer science and engineering. A finite state machine provides a basic mathematical computational model consisting of a definite set of states, transitions, and events. Finite state machines are flexible and easy to design, but as soon as the system grows in complexity, a reactivity/modularity trade-off problem arises and the approach becomes impractical. 
 
Decision trees~\cite{breiman1984classification}, the subsumption architecture~\cite{brooks1986robust}, and sequential behavior composition~\cite{burridge1999sequential} are widely-used approaches to decision making and task execution in robotics. A decision tree is an analytical decision support tool consisting of control structures and predicates located in the leaves and internal nodes, respectively, which map the possible outcomes of a series of choices. One motivation for their use is their ``white-box'' nature, i.e.\ decisions can be intuitively explained, they are simple to visualize, and may be easily implemented. However, decision trees lack robustness to noisy perceptual data, and their size rapidly increases in complex scenarios. The subsumption architecture relies on having a number of controllers in parallel which are ordered with different priorities so that each one is allowed to output both actuator commands and a binary value, signaling if the control of the robot is active or inactive. {In sequential behavior composition, the behavior is driven by local controllers. The state space is split into cells, corresponding to the basin of attraction of each controller. The task of each controller is to drive the system into the basin of attraction of another controller that is closer to the overall goal state.} 

Colledanchise and {\"O}gren~\cite{colledanchise2016behavior} have shown that \acp{bt} represent an elegant generalization of finite state machines, decision trees, the subsumption architecture, and sequential behavior composition. Several fixed-logic control models have emerged that extend the functionalities of \acp{bt} and attempt to overcome their limitations in highly dynamic environments. For instance, Conditional \acp{bt} \cite{giunchiglia2019conditional} extend traditional \acp{bt} to monitor the execution of an action considering logic pre- and postconditions. The work in~\cite{segura2017integration} proposes to cast an automated plan created by a hierarchical task network planner~\cite{nau2003shop2, kaelbling2011hierarchical} into executable \acp{bt}. In this way, \acp{bt} provide reactiveness and modularity, whereas the planner is responsible for the deliberative behavior of the robot. Along the same lines, other work~{\cite{rovida2017extended,neufeld2018hybrid, lan2019autonomous} explores different ways of synthesizing a \ac{bt} using a planner.} 
{Automatic synthesis is employed to produce \acp{bt} with guaranteed performance \cite{paxton2019representing} or safety \cite{tadewos2019automatic}}. Other approaches extend \acp{bt} by integrating models where domain knowledge can be learned automatically, for instance using {reinforcement learning~\cite{zhang2017modeling,zhu2019behavior}}, genetic programming~\cite{zhang2018learning}, or imitation learning~\cite{french2019learning}. Learning techniques can overcome the limitations of classical planners that require significant engineering effort~\cite{ilghami2005learning}. Finally, the work in~\cite{robertson2015building} attempts to map the practical solutions developed for action sequencing in real-time strategy games to robotic applications. 




Characteristics like human-readability, expressivity, modularity, and reusability make \ac{bt}-based techniques popular and unquestionably attractive. However, critical aspects that these approaches do not cover are the connection to the physical executive state and the possibility of ambiguities in the decision making. Their rigidity is not desirable for systems with strong perceptual constraints, like robots intended to interact with the environment and able to flexibly behave and react to unexpected events. To tackle this issue, some work exploits attentional mechanisms for visual task learning~\cite{borji2010online}, for cognitive control of humanoid robots~\cite{kawamura2008implementation}, and for flexible orchestration of cooperative tasks~\cite{caccavale2016flexible}. Inspired by the way humans monitor and orchestrate task execution~\cite{norman1986attention, cooper2006hierarchical}, the attentional framework in~\cite{caccavale2016flexible} loads tasks from a long-term memory and instantiates them in a working memory using a mechanism analogous to that used in Hierarchical Task Network planning~\cite{nau2003shop2}. Additionally, continuous sensory data are exploited to solve any ambiguities in the task plan and to quickly react to environmental changes. This attentional system has been effectively integrated in an imitation learning framework to learn, plan, and monitor complex robotic tasks~\cite{caccavale2017imitation, caccavale2019kinesthetic, saveriano2019symbolic, agostini2020manipulation}.

In this work, we take the best of these two worlds and propose \acp{rbt}, an executive framework that combines the human-readability, modularity, and reusability of \acp{bt} with the additional flexibility offered by attention-based cognitive architectures.  
 
\section{Reconfigurable Behavior Trees}
\label{sec:rbt}
\begin{table}[t!]
\caption{Types of BT nodes and their return status}
\resizebox{1\columnwidth}{!}{%
\begin{tabular}{|l|c|c|c|}
\hline
\textbf{Type} & \textbf{Symbol} & \textbf{Success} & \textbf{Failure}\\
 \hline \hline
\textit{Fallback/Selector} & $?$ & One child succeeds &  All children fail \\ \hline
\textit{Sequence} &      $\rightarrow$                  & All children succeed           & One child fails               \\ \hline
\textit{Parallel}  & $\rightrightarrows$                      & \textgreater$M$ children succeed &  \textgreater$N-M$ children fail \\ \hline
\textit{Decorator}   &   $\Diamond$                 & Custom                             & Custom                                           \\ \hline \hline
\textit{Action}      & $\hrectangle$                  & Upon completion                    & Impossible to complete                    \\ \hline
\textit{Condition} & \tikz \draw (1,1) ellipse (5pt and 2.5pt);                 & \texttt{True}     & \texttt{False}                                            \\ \hline
\end{tabular}}
\label{tab:btformulation}
\end{table}
\begin{figure}[t!]
  \centering
	\begin{tikzpicture}[transform shape, grow = down,
level distance=1cm,
level 1/.style={sibling distance=3cm},
level 2/.style={sibling distance=1.4cm}]
  \node [Sequence] {}
    child { node [Fallback, anchor=center] {}
      edge from parent [below] {}
      child { node [Condition, color=black, text=black] {object \\ picked}
      	edge from parent [below] {} }
      child { node [Action, color=black, text=black] {pick \\ object}
      	edge from parent [below] {} }
     }
    child { node [Fallback, anchor=center] {}
      edge from parent [below] {}
      child { node [Condition, color=black, text=black] {object \\ used}
      	edge from parent [below] {} }
      child { node [Action, color=black, text=black] {use \\ object}
      	edge from parent [below] {} }
     }
    child { node [Fallback, anchor=center] {}
      edge from parent [below] {}
      child { node [Condition, color=black, text=black] {object \\ placed}
      	edge from parent [below] {} }
      child { node [Action, color=black, text=black] {place \\ object}
      	edge from parent [below] {} }
     }
  ;
\end{tikzpicture}
	\vspace*{-2mm}
  \caption{A \ac{bt} representing a pick-use-place task. $\rightarrow$ is a Sequence node; $?$ are Fallback nodes. Ellipses define condition nodes and rectangles define action nodes.}
    \label{fig:bt}
\end{figure}
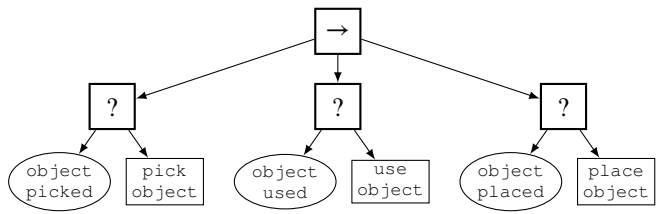

\begin{figure*}[t]
  \centering
  \usetikzlibrary{calc}
\usetikzlibrary{intersections}

\definecolor{dblue}{rgb}{0.2, 0.2, 0.6}

\begin{tikzpicture}[transform shape, grow = down,
level distance=1.2cm,
level 1/.style={sibling distance=4cm},
level 2/.style={sibling distance=7cm},
level 3/.style={sibling distance=3cm}]
  \node [Fallback, label=right:{\footnotesize \textcolor{dblue}{"rbt\_root"}}] {}
    child { node [Condition, inner sep=2, color=black, text=black] {goal \\ reached}}
    child { node [Sequence, label=right:{\footnotesize \textcolor{dblue}{"sequence\_1"}}] {}
      edge from parent [below] {}
      child { node [Fallback] {}
      edge from parent [below] {} 
      	child { node [Condition, inner sep=2, color=black, text=black] (D2) {blackboard \\ initialized}}
      	child { node [Action, inner sep=2, color=black, text=black] {initialize \\ blackboard}} }
      child { node [Parallel, label=right:{\footnotesize \textcolor{dblue}{"parallel\_1"}}] {}
      edge from parent [below] {}
      child { node [Action, inner sep=2, color=dgreen, text=black, label=left:{\footnotesize sensory data $\Rightarrow$}] {handle \\ priority}
      	edge from parent [below] {} } 
       	child { node [Fallback, label=right:{\footnotesize \textcolor{dblue}{"fallback\_1"}}] (D1) {}
      		 edge from parent [below] {} 
      		 child { node [Sequence] {}
      		 	edge from parent [below] {} 
      		 	 child { node [Condition, color=black, text=black] {priority \\ changed}
      		     edge from parent [below] {} }
      		 	 child { node [Action, color=black, text=black,  label=right:{\footnotesize LTM $\rightarrow$ WM}] {load \\ subtree}
      		     edge from parent [below] {} }
      		 }
      		 child { node [Action, color=black, text=black, label=right:{\footnotesize BT}]  {execute \\ subtree}
      		  edge from parent [below] {} }
      		  }
      		 }
      	}
     ;	
    \path[draw=orange,thick,name path=polygon] ($(D1.north east)+(2.7,0.3)$) -- ($(D1.north east)+(2.7,-3.3)$) -- ($(D1.north east)+(-4.5,-3.3)$) -- ($(D1.north east)+(-4.5,-2.3)$) -- ($(D1.north east)+(-1,0.3)$) -- cycle;
   \path[draw=dblue,thick,name path=polygon] ($(D2.south west)+(-.5,-0.3)$) -- ($(D2.south west)+(4.9,-0.3)$) -- ($(D2.south west)+(4.9,1.)$) -- ($(D2.south west)+(8.2,2.2)$) -- ($(D2.south west)+(8.2,4.5)$) -- ($(D2.south west)+(-0.5,4.5)$) -- cycle;
\end{tikzpicture}
  \caption{The generic RBT. The branch of the tree surrounded by the blue polygon allows execution to be terminated after a global goal is reached and is the same for RBT and BT. The green action node is the \textit{Emphasizer} that modifies the priority of each subtree considering the environmental stimuli. The branch of the tree surrounded by the orange polygon is dynamically allocated/deallocated by the \textit{Instantiator} every time the subtree priority order changes. The action node \texttt{execute subtree} contains small \acp{bt} like the one in Fig.~\ref{fig:bt}. The blue labels are the node names used in Listing~\ref{lst:jsonschema}.}
    \label{fig:rbt-overview}
\end{figure*}
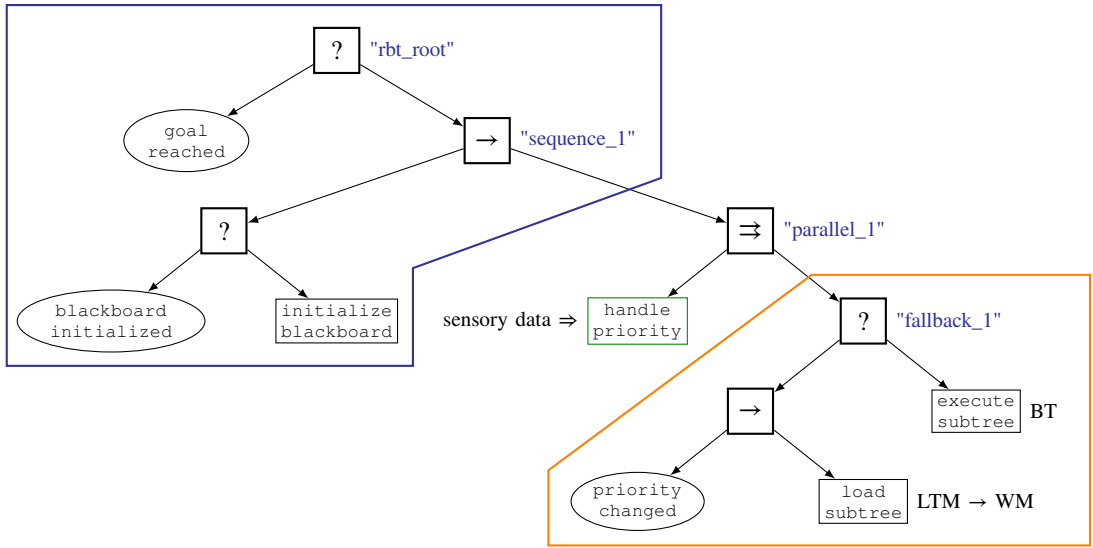

\subsection{Behavior Trees}\label{subsec:bt}
A \ac{bt} is a graphical model language to control {the actions (or behaviors)} of an autonomous agent and execute a task. 
A BT consists of a tree structure containing a combination of the six types of nodes shown in Table~\ref{tab:btformulation}. These types are divided into two categories: \textit{control flow} and \textit{execution} nodes. The four types of control flow nodes are Sequence, Fallback/Selector, Decorators, and Parallel nodes. The two types of execution nodes are Condition and Action nodes. Each type of node returns a \textit{running} state during the execution and \textit{success} or \textit{failure} after the execution. The execution of a \ac{bt} is possible by periodically traversing the tree from the root node to all child nodes from left to right. The traversing mechanism is periodically activated by sending a signal called ``tick''. Each child node responds to this signal according to its own type and to the return state of the other nodes. Table~\ref{tab:btformulation} describes the behavior of each type of node for the Success and Failure cases. The running state will behave in a similar fashion.

A minimal example of a \ac{bt} applied to a pick-use-place object task is shown in Fig.~\ref{fig:bt}. The root being a Sequence node, the BT is executed sequentially from left to right. If the condition \texttt{object picked} has not been fulfilled, i.e. the node returns \texttt{False}, the action node \texttt{pick object} enters the \textit{Running} state. The action node returns success upon successful completion of the pick action. As a consequence, the first (far left) Fallback node also returns success and the traversal mechanism enters the second (middle) Fallback node. This procedure is repeated until all the Fallback nodes return success, indicating that the task has been successfully completed.


\subsection{RBT components}
The generic \ac{rbt}, depicted in Fig.~\ref{fig:rbt-overview}, is a \ac{bt} enriched with extra functionalities that permits the continuous monitoring of environmental stimuli and the dynamic reconfiguration of the tree to execute. Interestingly, those functionalities are implemented using the same six types of nodes considered in the standard \acp{bt} (see Table~\ref{tab:btformulation}), leaving unaffected the design simplicity typical of \acp{bt}. Traversing the RBT from the root (Fallback) node, we first check if the end goal of the task is fulfilled. If not, we check if the blackboard is initialized and eventually run  \texttt{initialize blackboard}. The blackboard is a mechanism used in \acp{bt} to store and update runtime variables and to make them accessible by each node in the tree. In the \ac{rbt} framework, the blackboard contains the logical pre- and postconditions used to regulate the task execution and to determine when the task goal is fulfilled, the sensed information used to set the priorities of each subtree, and the current value of the subtrees priorities. The blackboard is dynamically updated and greatly simplifies the communication between nodes by handling the concurrent access in a transparent and thread safe manner. We would like to point out that this part of the \ac{rbt}, surrounded by a blue polygon in Fig.~\ref{fig:rbt-overview},  allows execution to be terminated after a global goal is reached. A similar branch has to be introduced in the standard BT to successfully terminate the task and therefore is does not introduce extra nodes in the RBT. Once the blackboard is initialized the right branch of the RBT is traversed and two parallel processes start. On one side, sensory data are processed to determine the priority of the $S$ available subtrees. On the other side, the most emphasized subtree is instantiated and executed asynchronously with respect to the perceptual input. The instantiation mechanism is dynamic and allows for flexible task orchestration.  

Compared to a BT, the RBT has the following extra components:
\begin{enumerate}
\item A \textit{Long-Term Memory (LTM)} and a \textit{Short-Term} or \textit{Working Memory (WM)} that are typical of attention-based control frameworks~\cite{caccavale2016flexible}. 
\item A priority handler, namely the \textit{Emphasizer}, that computes the highest-priority task considering the sensed information and logical pre- and postconditions.
\item An \textit{Instantiator} that accesses the LTM and casts the subtask into the corresponding subtree loaded in the WM. The Instantiator enables the \ac{rbt} reconfiguration capabilities by dynamically loading and instantiating the subtree with higher priority. 
\end{enumerate}
The distinctive components of \acp{rbt} are detailed in the rest of this section.


\subsection{LTM and WM}
The LTM can be considered a database that contains all subtasks that the robot is able to execute. A typical subtask is the pick-use-place object task described in Sec.~\ref{subsec:bt}. In order to store and retrieve subtasks from the LTM, we propose a common representation of the $4$ control flow nodes in Table~\ref{tab:btformulation}. The generic control flow node $\mathcal{B}$ is represented as a quadruple
\begin{equation}
\mathcal{B}=(l,t,c,p)
\label{eq:node_tuple}
\end{equation} 
where $l$ is the unique name (label) of the node, $t$ is one of the types in Table~\ref{tab:btformulation}, $c$ is a list of children, and $p$ is a list of \textit{parameters} like the priority value or pre- and postconditions. 

In principle, it is possible to represent also the $2$ execution nodes in Table~\ref{tab:btformulation} as the quadruple defined in~\eqref{eq:node_tuple}. However, we found a more convenient way of exploiting the fact that execution nodes correspond to leaves in the BT. In more detail, action nodes are specified only in the children list of the parent node, while the condition nodes are used to represent the pre- and postconditions that are listed in the  parameter list $p$. The successful execution of an action also changes the state of the relative postcondition, while the preconditions of an action are changed by other nodes in the tree.

Following this representation, the LTM can be conveniently organized in JSON schemas. As shown in Listing~\ref{lst:jsonschema}, the root of a tree is identified by the key word \texttt{root} in its name (\texttt{rbt\_root}). Actions are simply listed as children of a node and are identified by the string \texttt{A(action\_name)}. For preconditions, we use the notation \texttt{C\_ij} where \texttt{i} is the child number and \texttt{j} indicates the $j$-th condition. Postconditions (or goals) are identified by \texttt{G\_ij} where \texttt{i} is the child number and \texttt{j} indicates the $j$-th condition. It is worth noticing that the described representation contains all the information needed to instantiate an executable BT and that no further JSON schemas are needed to describe the leaves of the BT. 


 \begin{lstlisting}[linewidth=\columnwidth,label=lst:jsonschema, basicstyle=\small\ttfamily, float, caption=JSON schemas representing the generic RBT in Fig.~\ref{fig:rbt-overview}.]
{
  "name":"rbt_root",
  "type":"fallback",
  "children": ["sequence_1"],
  "params": ["G_11", "goal reached"]
},  
{
  "name":"sequence_1",
  "type":"sequence",
  "children": ["A(initialize blackboard)",
                "parallel_1"],
  "params": ["G_11",
             "blackboard initialized"]
},
{
  "name":"parallel_1",
  "type":"parallel",
  "children": ["A(handle priority)",
                "fallback_1"],
  "params": [""]
}
{
  "name":"fallback_1",
  "type":"fallback",
  "children": ["A(load subtree)",
                "A(execute subtree)"],
  "params": ["C_11", "priority changed"]
}
\end{lstlisting}

The \textit{Instantiator} is responsible for loading the task from the LTM and creating an instance of the BT to execute in the WM. This procedure is summarized in Algorithm~\ref{alg:allocate-tree}. Given the task name (root of the BT), the Instantiator first loads the JSON schemas describing the task (line $2$ in Algorithm~\ref{alg:allocate-tree}). Starting from the root, the BT is built by iteratively expanding the nodes until the leaves are reached (lines $4$ to $19$). The current JSON schema is converted into the BT node specified in the \texttt{type} field and attached to the current tree (line $5$). Pre- and postcondtions are handled using a modified version of the Planning and Acting PA-BT approach in~\cite{colledanchise2018behavior} that allows multiple postconditions. In this approach, a postcondition is transformed into a Condition node (line $7$) that is connected to the rest of the tree via a Fallback node ($\mathcal{T}_{fal}$). In this way, execution ends when the postcondition becomes \texttt{True}. The case of a single postcondition is handled in lines $8$--$9$. In case of multiple postconditions, a Sequence node is created with all the postconditions attached (line $11$). In this way, the postconditions are sequentially verified. The Sequence node containing all the postconditions is then attached to $\mathcal{T}_{fal}$ (line $12$). In both cases the generated Fallback node is attached to the current tree (line $18$). Lines $14$--$17$ handle Action nodes with associated preconditions. As for the postconditions, the preconditions are considered as Condition nodes (line $16$). Action and Condition nodes are then connected to a Sequence node that is attached to the current tree (lines $17$--$18$). In this way, an Action is executed only if all the preconditions are \texttt{True}. As a final note, the functions \textsc{sequenceNode} and \textsc{fallbackNode} in Algorithm~\ref{alg:allocate-tree} return an empty subtree if the input is empty, while \textsc{sequenceNode}($\{\}$, $\mathcal{A}$) returns the Action nodes $\mathcal{A}$.

\begin{algorithm}[t]
  \begin{algorithmic}[1]
    \Function{instantiateSubTree}{$l$} \Comment{$l$: subtask name}
		\State \texttt{schemaList} $ \leftarrow$ \textsc{getTaskFromLTM}($l$) 
    \State $\mathcal{T} \leftarrow \{\}$ \Comment{empty BT}
    \For{\texttt{schema} in \texttt{schemaList}}
      \State $\mathcal{T} \leftarrow$ \textsc{schemaToNode}($\mathcal{T}$, \texttt{schema})
      \State \texttt{postC} $ \leftarrow$ \textsc{getPostConditions}(\texttt{schema}) 
      \State $\mathcal{C} \leftarrow$ \textsc{conditionNodes}(\texttt{postC})
      \If{$\mathcal{C}$\texttt{.length} $== 1$}
        \State $\mathcal{T}_{fal} \leftarrow$ \textsc{fallbackNode}(\texttt{$\mathcal{C}$})
      \Else 
        \State $\mathcal{T}_{seq}$ $\leftarrow$ \textsc{sequenceNode}($\mathcal{C}$)
        \State $\mathcal{T}_{fal} \leftarrow$ \textsc{attachSubTree}($\mathcal{T}_{seq}$)
      \EndIf
      \State \texttt{a}, \texttt{preC}  $ \leftarrow$ \textsc{getActions}(\texttt{schema}) 
      \State $\mathcal{A} \leftarrow$ \textsc{actionNodes}(\texttt{a})
      \State $\mathcal{C} \leftarrow$ \textsc{conditionNodes}(\texttt{preC})
      \State $\mathcal{T}_{seq}$ $\leftarrow$ \textsc{sequenceNode}($\mathcal{C}$, $\mathcal{A}$)
      \State $\mathcal{T} \leftarrow $ \textsc{attachSubTree}($\mathcal{T}_{fal}$, $\mathcal{T}_{seq}$)  
    \EndFor
    \State \Return{$\mathcal{T}$}
    \EndFunction
  \end{algorithmic}
  \caption{Load and instantiate a BT}
  \label{alg:allocate-tree}
\end{algorithm}

\subsection{Subtree priority}
\label{subsec:priority}
The modularity of standard BT allows a complex task (tree) to be decomposed into subtasks (subtrees). For instance, a stacking task can be decomposed as a combination of pick and place subtasks. However, in standard BT, the execution order of each subtask is predefined. Changing the execution order depending on discrete values is possible, but requires extra branches in the BT. 
Changing the execution order by monitoring continuous values like sensory data is typically not supported.  

In contrast, \acp{rbt} exploit logical pre- and postconditions, as well as continuous sensory data, to monitor the task execution. As already mentioned, a complex task is divided into subtrees. We assign pre- and postconditions to the root of each subtree. Therefore, a specific subtree is \emph{active} if all the preconditions are satisfied while the postconditions are not. A subtree correctly terminates by setting the postconditions. At each tick, the Emphasizer accesses the blackboard and looks for active subtrees. An execution conflict occurs every time more than one active subtree exists. In this case, we exploit a priority-based mechanism to dynamically decide the subtask to execute. We define the priority of a subtree as the runtime prominence for the execution of a specific subtask. The priority is a real value, normalized between $0$ and $1$, which tells the Instantiator which subtask to load and transform into an executable BT.
In this work, the priority $\epsilon$ is defined as
\begin{equation}
\label{eq:emph}
\epsilon(\theta)=
\begin{cases}
  1 & \textrm{if } \theta \leq \theta_{\min} \\
 \dfrac{\theta-\theta_{\max}}{\theta_{\min} - \theta_{\max}} & \textrm{if } \theta_{\min} < \theta < \theta_{\max} \\ 
 0 & \textrm{if }  \theta \geq \theta_{\max}
 \end{cases}
\end{equation}
where $\theta$ is a continuous value coming from sensory data, and the thresholds $\theta_{\min}$ and $\theta_{\max}$ are tunable parameters. In this work, $\theta$ represents the inverse of the distance from objects to manipulate, but other choices are possible.


\subsection{Task execution and monitoring}
\label{sec:task_execution}
The generic \ac{rbt}, as the one depicted in Fig.~\ref{fig:rbt-overview}, is a goal-oriented tree that successfully terminates only if a certain goal is achieved (the \texttt{goal reached} condition becomes \texttt{True}). In our implementation, the goal of the RBT is achieved if the postconditions of all the subtrees are satisfied.    
As already mentioned, the execution of the RBT is achieved by periodically traversing the tree top to bottom and left to right (tick function). At each tick, the \texttt{goal reached} condition is tested. If  \texttt{goal reached} is \texttt{False}, we check that the blackboard is initialized and then enter a Parallel node. Executing the Emphasizer and the subtree in parallel is convenient because sensory data and task execution are, in general, asynchronous processes. The Parallel node is designed to successfully terminate if both children are successfully executed. Since the Action node \texttt{handle priority} is always in the running state, the Parallel node cannot terminate. This implies that the RBT successfully terminates if and only if the goal is reached. 

If new sensory data arrive or a subtree postcondition changes, the Emphasizer recomputes the priority of the subtrees and sets the \texttt{priority changed} flag. This triggers the Instantiator that preempts the current execution, deallocates the current subtree, and instantiates the subtree with highest priority. As already mentioned, this branch of the tree---the branch inside the orange polygon in Fig.~\ref{fig:rbt-overview}---is dynamically allocated at each tick. The dynamic allocation is needed for the correct execution of the task. To better understand this point, consider what happens if \texttt{execute subtree} returns \texttt{True}. In this case, the active subtask has been successfully executed and the Fallback node (\texttt{fallback\_1}) also returns \texttt{True} (see Table~\ref{tab:btformulation}). With a static branch, the tick would not enter \texttt{fallback\_1} anymore, letting the remaining active subtasks unexecuted. With a dynamic branch, instead, the return state of \texttt{fallback\_1} is reset and the active subtask with highest priority is correctly instantiated and executed.

\section{Evaluation}\label{sec:evaluation}
\begin{figure}[t]
	\centering
	\subfigure[Possible initial configuration.]{\includegraphics[width=0.48\columnwidth]{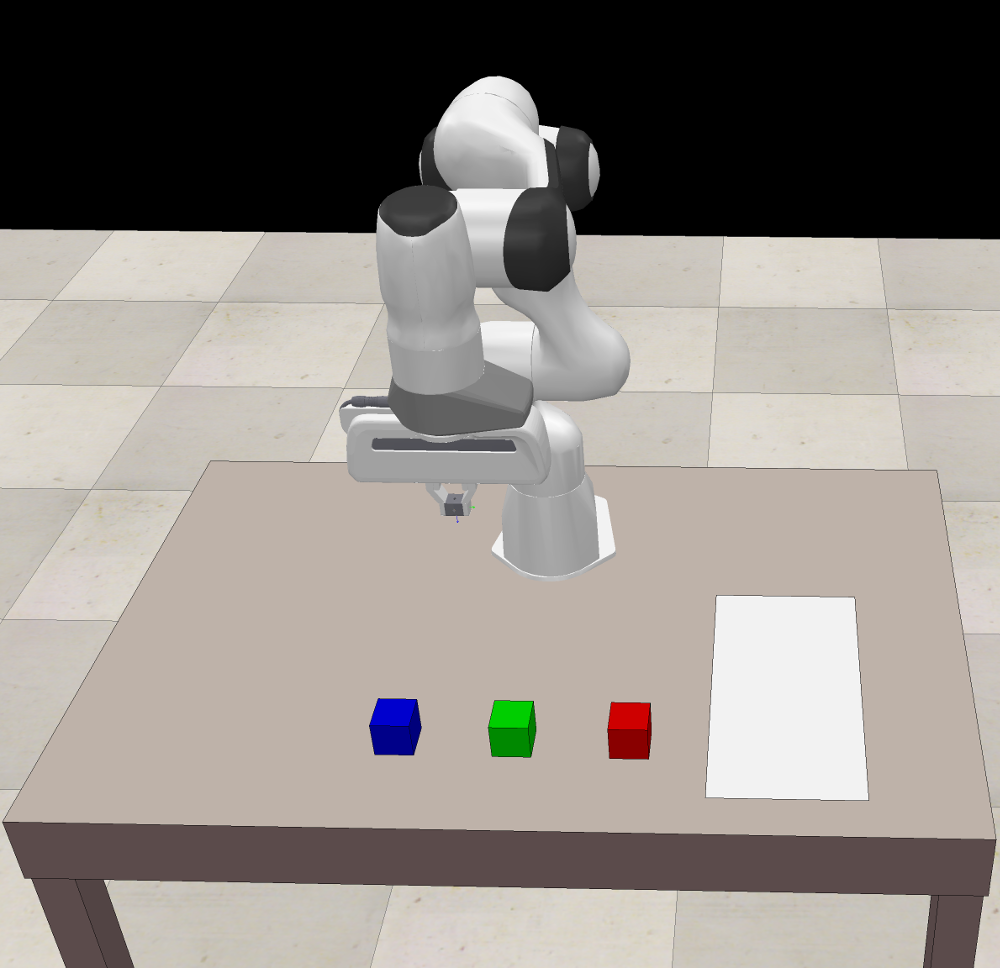}}
	\subfigure[Desired final configuration.]{\includegraphics[width=0.48\columnwidth]{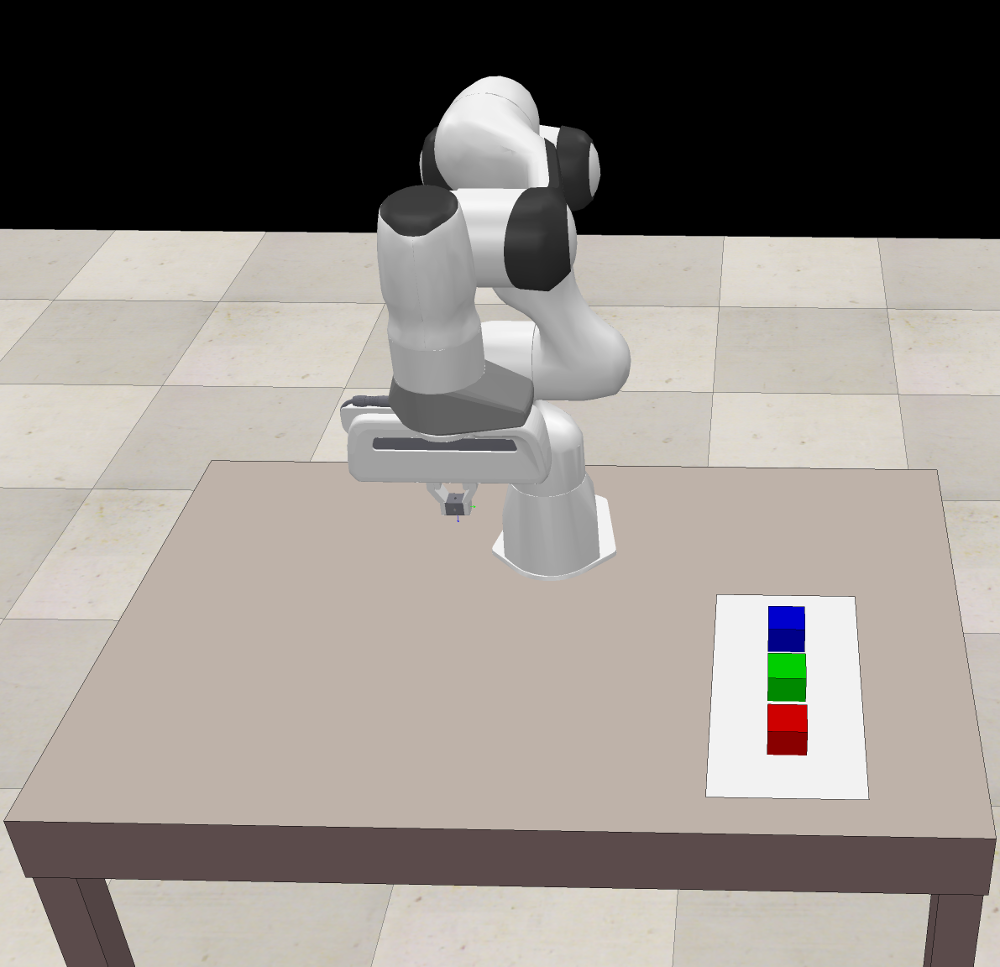}}
  \caption{The sorting task where the robot has to pick $3$ boxes from the table and place them at a specific location in the white storage area.}
  \label{fig:setup}
\end{figure}
\begin{figure}[t!]
  \centering
	\begin{tikzpicture}[transform shape, grow = down,
level distance=1cm,
level 1/.style={dashed, sibling distance=2cm},
level 2/.style={solid, sibling distance=2.5cm},
level 3/.style={sibling distance=1.5cm}]
  \node [Sequence,dashed] {}
  child { node [Condition, color=black, text=black] {C\_11}
          edge from parent [below] {} }
  child { node [Condition, color=black, text=black] {C\_12}
          edge from parent [below] {} }
  child { node [Fallback, solid, label={[xshift=1.5cm, yshift=0.5cm]\texttt{sort box}}] {}
  			edge from parent [below] {}
  child { node [Condition, thick, color=black, text=black] {box \\ placed}
          edge from parent [below, thick] {} }
  child { node [Sequence, anchor=center] {}
    child { node [Fallback, anchor=center] {}
      edge from parent [below] {}
      child { node [Condition, color=black, text=black] {box \\ picked}
      	edge from parent [below] {} }
      child { node [Action, color=black, text=black] {pick \\ box}
      	edge from parent [below] {} }
     }
     child { node [Action, color=black, text=black] {place \\ box}
      	edge from parent [below] {} }
   }
   }
  ;
\end{tikzpicture}
	\vspace*{-2mm}
  \caption{The BT used to pick a box from the table and place it in the storage area. This BT is compactly represented by the \texttt{execute subtree} node in the RBT of Fig.~\ref{fig:rbt-overview}. The RBT uses the dashed nodes to impose the preconditions in case study~1, while in case study~2 they are omitted.}
    \label{fig:bt-pick-place}
\end{figure}
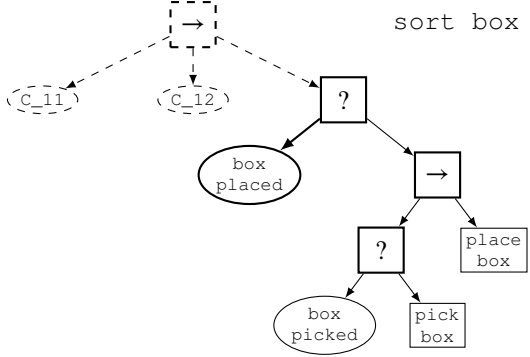
We evaluate \acp{rbt} in the sorting task shown in Fig.~\ref{fig:setup} where the robot has to pick $3$ colored boxes (red, blue, and green) from a table (Fig.~\ref{fig:setup}(a)) and place them at specific locations in a storage area (Fig.~\ref{fig:setup}(b)). The scenario is simulated in CoppeliaSim~\cite{coppeliaSim} using the Panda robot model provided by Gaz et al.~\cite{gaz2019dynamic}. We consider two different case studies and compare the performance of \acp{rbt} and \acp{bt} in terms of execution time and tree complexity---the number of nodes in the tree. {Note that the execution time is computed by summing up the tick times until the goal is reached and without considering the time spent by the robot to perform the commanded actions.} \acp{bt} are implemented using the open-source Python library \texttt{py\_tree}~\cite{pytrees_doc}. \acp{rbt} are also implemented in Python exploiting the basic \ac{bt} nodes provided by \texttt{py\_tree}.

Figure~\ref{fig:bt-pick-place} shows the BT used to plan box sorting subtasks. The solid nodes are common between \acp{bt} and \acp{rbt}, while the dashed nodes are exploited by \acp{rbt} to enforce a certain execution order specified by a set of preconditions \{\texttt{C\_11}, \texttt{C\_12}\}. As discussed in Sec.~\ref{sec:rbt}, \acp{rbt} dynamically attach the subtree in Fig.~\ref{fig:bt-pick-place} to the \textit{static} tree in Fig.~\ref{fig:rbt-overview}. Moreover, the standard BT is endowed with $6$ extra nodes (contained in the blue polygon in Fig.~\ref{fig:rbt-overview}) to monitor the execution of the task and successfully terminate after a goal is reached. The task is fulfilled when all the boxes are sorted in the storage area, as indicated by the task goal \texttt{G\_11 = b\_box placed $\land$ g\_box placed $\land$ r\_box placed}. The thresholds used to compute the priority in~\eqref{eq:emph} are empirically set to $\theta_{\min}=0.05\,$m (the length of the box side) and to $\theta_{\max}=1\,$m (the maximum distance that still allows grasping a box). We compare RBT and BT in terms of number of nodes and execution time measured assuming that action nodes directly return \texttt{True} without entering the running state.



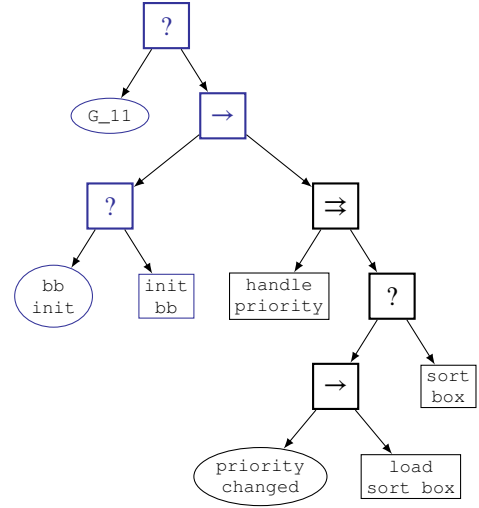
\begin{figure}[t!]
  \centering
  \usetikzlibrary{calc}
\usetikzlibrary{intersections}

\definecolor{dblue}{rgb}{0.2, 0.2, 0.6}

\begin{tikzpicture}[transform shape, grow = down,
level distance=1.2cm,
level 1/.style={sibling distance=1.5cm},
level 2/.style={sibling distance=3cm},
level 3/.style={sibling distance=1.5cm},
level 5/.style={sibling distance=2.cm}]
  \node [Fallback, color=dblue] {}
    child { node [Condition, inner sep=2, color=dblue, text=black] {G\_11}}
    child { node [Sequence, color=dblue] {}
      edge from parent [below] {}
      child { node [Fallback, color=dblue] {}
      edge from parent [below] {} 
      	child { node [Condition, inner sep=2, color=dblue, text=black] {bb \\ init}}
      	child { node [Action, inner sep=2, color=dblue, text=black] {init\\ bb}} }
      child { node [Parallel] {}
      edge from parent [below] {}
      child { node [Action, inner sep=2, color=black, text=black] {handle \\ priority}
      	edge from parent [below] {} } 
       	child { node [Fallback] {}
      		 edge from parent [below] {} 
      		 child { node [Sequence] {}
      		 	edge from parent [below] {} 
      		 	 child { node [Condition, color=black, text=black] {priority \\ changed}
      		     edge from parent [below] {} }
      		 	 child { node [Action, color=black, text=black] {load \\ sort box}
      		     edge from parent [below] {} }
      		 }
      		 child { node [Action, color=black, text=black]  {sort \\ box}
      		  edge from parent [below] {} }
      		  }
      		 }
      	}
     ;	
\end{tikzpicture}
  \caption{The RBT used to plan the sorting task. In case study 1, \texttt{sort box} is subject to preconditions to constrain the execution. The blue nodes are common to \acp{bt} and \acp{rbt}.}
    \label{fig:rbt-sorting}
\end{figure}
%
%
\begin{table}[t]
\caption{Comparison between RBT and standard BT.}
\begin{center}
\resizebox{0.8\columnwidth}{!}{%
\begin{tabular}{|c|c|c|c|}
\hline
 Method & Case & \# Nodes & Execution Time (ms) \\ \hline \hline
RBT & 1 & 19 - 22   &  507.04\\ \hline
BT & 1 & 27   &  503.24 \\ \hline \hline
RBT & 2 & 19 &   505.17\\ \hline
BT & 2 &  151 &  819.78\\ \hline 
\end{tabular}}
\end{center}
\label{tab:tabev1}
\end{table}

\subsubsection{Case study 1}
The goal of this experiment is to compare \acp{rbt} and \acp{bt} in a situation that favours the BT, i.e. when the task plan is rigid, no ambiguities are possible, and no external disturbances occur. In this case, we assume that the boxes need to be sorted in a specific order: First we pick and place the blue box \texttt{b\_box}, then the green box \texttt{g\_box}, and finally the red box \texttt{r\_box}. This sorting order has been arbitrarily decided and does not affect the obtained results. The sorting task successfully terminates when the task goal \texttt{G\_11} is reached. 
\begin{figure}[t!]
  \centering
	\subfigure[]{\usetikzlibrary{calc}
\usetikzlibrary{intersections}

\definecolor{dblue}{rgb}{0.2, 0.2, 0.6}

\begin{tikzpicture}[transform shape, grow = down,
level distance=1.2cm,
level 1/.style={sibling distance=1.5cm}]
  \node [Fallback, color=dblue] {}
    child { node [Condition, inner sep=2, color=dblue, text=black] {G\_11}}
    child { node [Sequence, color=dblue] {}
      edge from parent [below] {}
      child { node [Fallback, color=dblue] {}
      edge from parent [below] {} 
      	child { node [Condition, inner sep=2, color=dblue, text=black] {bb \\ init}}
      	child { node [Action, inner sep=2, color=dblue, text=black] {init \\ bb}} }
			child { node [Action, inner sep=2, color=black, text=black] {sort \\ b\_box }}  
			child { node [Action, inner sep=2, color=black, text=black] {sort \\ g\_box }}
			child { node [Action, inner sep=2, color=black, text=black] {sort \\ r\_box }}     	
      	}
     ;	
\end{tikzpicture}}
	\subfigure[]{\usetikzlibrary{calc}
\usetikzlibrary{intersections}

\definecolor{dblue}{rgb}{0.2, 0.2, 0.6}

\begin{tikzpicture}[transform shape, grow = down,
level distance=1.2cm,
level 1/.style={sibling distance=1.5cm},
level 2/.style={sibling distance=3cm},
level 3/.style={sibling distance=1.5cm}]
  \node [Fallback, color=dblue] {}
    child { node [Condition, inner sep=2, color=dblue, text=black] {G\_11}}
    child { node [Sequence, color=dblue] {}
      edge from parent [below] {}
      child { node [Fallback, color=dblue] {}
      edge from parent [below] {} 
      	child { node [Condition, inner sep=2, color=dblue, text=black] {bb \\ init}}
      	child { node [Action, inner sep=2, color=dblue, text=black] {init \\ bb}} }
      child { node [Fallback, color=black] {}
      	edge from parent [below] {} 
			child { node [Sequence, color=black] {}
			edge from parent [below] {} 
				child { node [Condition, inner sep=2, color=black, text=black] {$d_b \leq d_g$}}
				child { node [Condition, inner sep=2, color=black, text=black] {$d_g \leq d_r$}}
				child { node [Action, inner sep=2, color=black, text=black] {sort \\ b\_box}}
				child { node [Action, inner sep=2, color=black, text=black] {sort \\ g\_box}}
				child { node [Action, inner sep=2, color=black, text=black] {sort \\ r\_box}}
      }
      	child { node [shape=rectangle, text=black, font=\Large] {$\ldots$}}
      }
      }
    ;	
\end{tikzpicture}}
  \caption{The BT used to plan the sorting task of case study 1 (a) and case study 2 (b). The blue nodes are common to \acp{bt} and \acp{rbt}. The variable $d_i$, $i=b,r,g$, indicates the distance between the \texttt{i\_box} and the robot. Due to the limited space, we only show a partial \ac{bt} (b). In particular, the stump containing the black sequence node and its children ($24$ nodes in total) handles the cases $d_b \leq d_g$ and $d_g \leq d_r$. Five similar stumps are needed to handle all possible combinations of box distances and are compactly indicated here by the symbol $\cdots$.}
    \label{fig:bt-pick-place-cases}
\end{figure}
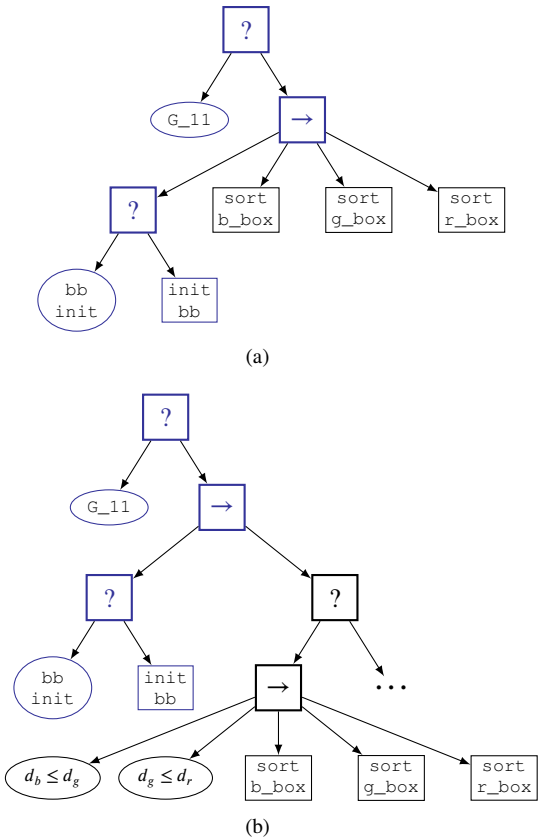
The BT used to perform the sorting task is shown in Fig.~\ref{fig:bt-pick-place-cases}(a), where the $3$ Action nodes \texttt{sort} \texttt{\textsl{box\_name}} compactly represent the BT in Fig.~\ref{fig:bt-pick-place} (only the solid nodes are considered). As listed in Table~\ref{tab:tabev1}, the BT of Fig.~\ref{fig:bt-pick-place-cases}(a) has $27$ nodes. The RBT used to plan this task is shown in Fig.~\ref{fig:rbt-sorting} and is almost identical in the two case studies. The only difference is in the \texttt{sort box} Action node. In case study 1, we exploit preconditions (dashed nodes in Fig.~\ref{fig:bt-pick-place}) to impose a certain execution order. More specifically, the Action node \texttt{sort box} is allocated by the \textit{Instantiator} that, at runtime, dynamically instantiates a specialized version of the sorting task where the generic box is replaced by the one with highest priority. As described in Sec.~\ref{subsec:priority}, the priority of a subtask depends on logical precondition and continuous stimuli. In this case, external stimuli play no role and the execution order is fully determined by the preconditions. In particular, \texttt{sort b\_box} has no preconditions and is the first to be executed. \texttt{sort g\_box} has \texttt{C\_11 = b\_box placed} as precondition, while \texttt{sort r\_box} has \texttt{C\_11 = b\_box placed} and \texttt{C\_12 = g\_box placed} as preconditions. This guarantees that the sorting task is executed in the desired sequential order. As listed in Table~\ref{tab:tabev1}, the RBT has a variable number of nodes ($19$ to $22$). This is because the \texttt{sort} node has a variable number of preconditions for each box. Even if in case study 1 the \ac{rbt} does not show its full potential, we still have a reduction of the number of nodes in the tree ($22$ nodes in RBT in the worst case, $27$ in the \ac{bt}). The time to execute the \ac{bt} is slightly smaller in this case, but the difference is negligible (less than $4\,$ms).


\subsubsection{Case study 2}
This experiment is designed to show the benefits of the priority-based task execution introduced by \acp{rbt}. The scenario is the same as in case study~1 but here the boxes are sorted without a predefined order. {In principle, there are $6$ possibilities (plans) to pick the $3$ boxes in Fig.~\ref{fig:setup}(a) and place them in the goal configuration shown in Fig.~\ref{fig:setup}(b). \texttt{Sort b\_box, r\_box, g\_box} and \texttt{sort b\_box, g\_box, r\_box} are $2$ of the $6$ of feasible plans. We use the robot-box distances to determine which plan the robot executes. In particular, we always sort first the box closest to the gripper.}. The \ac{rbt} used to plan this task is the same as in case study 1, except for the \texttt{sort box} Action node that has no general preconditions. Therefore, the \ac{rbt} always has $19$ nodes (see Table~\ref{tab:tabev1}). At each time, all the boxes on the table are eligible to be sorted. These execution conflicts are managed in \acp{rbt} using the task priorities computed with~\eqref{eq:emph}. Hence, the \textit{Instantiator} is free to allocate the subtree with highest priority, i.e. to start sorting the closest box. Once a box, for instance the blue one, is placed in the storage area, the postcondition \texttt{b\_box placed} becomes \texttt{True} and the \textit{Emphasizer} removes \texttt{sort b\_box} from the list of active nodes (see Sec.~\ref{subsec:priority}). This implies that \texttt{sort b\_box} is not instantiated in future ticks, letting the robot sort the other boxes and successfully complete the task. 
In order to reach a similar level of flexibility with standard \acp{bt}, we need {many more nodes to represent the $6$ feasible plans in one tree and a more complicated control flow logic to determine the plan to execute based on robot-box distances}. A possible solution that requires $151$ nodes is sketched in Fig.~\ref{fig:bt-pick-place}(b). In this case, \acp{rbt} require $\approx 85\,$\% less nodes, which clearly makes the RBT easier to visualize, and a reduction of the execution time of $\approx 38\,$\%.

\section{Conclusion and Future Work}
\label{sec:conclusion}
We introduced Reconfigurable Behavior Trees, a novel framework that combines high-level decision making and continuous monitoring and control features. By combining the expressivity and modularity of standard behavior trees with the flexibility of attentional execution monitoring, \acp{rbt} allow the AI agent to perform actions in a robust and versatile way, while being capable of adjusting its behavior based on continuous input from the environment. The proposed framework has been tested in a sorting task and compared with standard \acp{bt} in terms of tree complexity and computation time. The evaluation shows that \acp{rbt} outperform standard \acp{bt}, especially when the task plan is not rigidly defined and ambiguities in the execution need to be solved.

{This paper focused on theoretical description of \acp{rbt} and comparisons with traditional \acp{bt}. In \cite{saveriano2020combining}, we combined \acp{rbt} with a reactive motion generation approach to correct the robot trajectory on-the-fly without a computationally expensive motion re-planning step. The preliminary results shown in this paper and in \cite{saveriano2020combining} are promising and suggest that \acp{rbt} are better suited to dynamic and uncertain environments than \acp{bt}. Therefore, we expect that \acp{rbt} can significantly contribute to the deployment of social robotics applications. This clearly requires further development of the proposed framework. Among others, the \ac{rbt} framework has to be tested in human-robot interaction scenarios. Uncertainty introduced by the human in the loop can be estimated using, for instance, neural networks \cite{kabir2018neural} and can explicitly be considered in the decision process.}

\balance

\bibliographystyle{IEEEtran}
\bibliography{mybib}
\end{document}